\pdfoutput=1 
\documentclass{article}
\usepackage{ijcai22}

\usepackage{times}
\usepackage{soul}
\usepackage{url}
\usepackage[hidelinks]{hyperref}
\usepackage[utf8]{inputenc}
\usepackage[small]{caption}
\usepackage{amsfonts}
\usepackage{multirow}
\usepackage{multicol}
\usepackage{graphicx}
\usepackage{amsmath}
\usepackage{amsthm}
\usepackage{booktabs}
\usepackage{algorithm}
\usepackage{algorithmic}
\title{Deep Video Harmonization with Color Mapping Consistency}

\author{
Xinyuan Lu$^{1*}$
\and
Shengyuan Huang$^{1}$\thanks{Both authors contributed equally to this research.}\and
Li Niu$^{1\dag}$\and
Wenyan Cong$^1$\and
Liqing Zhang$^{1}$\thanks{Corresponding Authors.}
\affiliations
$^1$MoE Key Lab of Artificial Intelligence, Department of Computer Science and Engineering\\
Shanghai Jiaotong University, Shanghai, China
\emails
\{lxy9807, huangshengyuan, ustcnewly, plcwyam17320\}@sjtu.edu.cn, zhang-lq@cs.sjtu.edu.cn.
}

\begin{document}

\maketitle

\begin{abstract}
Video harmonization aims to adjust the foreground of a composite video to make it compatible with the background. So far, video harmonization has only received limited attention and there is no public dataset for video harmonization. In this work, we construct a new video harmonization dataset HYouTube by adjusting the foreground of real videos to create synthetic composite videos. Moreover, we consider the temporal consistency in video harmonization task. Unlike previous works which establish the spatial correspondence, we design a novel framework based on the assumption of color mapping consistency, which leverages the color mapping of neighboring frames to refine the current frame. Extensive experiments on our HYouTube dataset prove the effectiveness of our proposed framework. Our dataset and code are available at https://github.com/bcmi/Video-Harmonization-Dataset-HYouTube. 
\end{abstract}

\section{Introduction}
Given two different videos, video composition aims to generate a composite video by combining the foreground of one video with the background of another video. However, composite videos are usually not realistic enough due to the appearance (\emph{e.g.}, illumination, color) incompatibility between foreground and background, which is caused by distinctive capture conditions (\emph{e.g.}, season, weather, time of the day) of foreground and background \cite{2020DoveNet,2021Bargainnet}. To address this issue, video harmonization \cite{2019Temporally} has been proposed to adjust the foreground appearance to make it compatible with the background, resulting in a more realistic composite video.

\begin{figure}[t]
    \centering
    \includegraphics[width=0.4\textwidth]{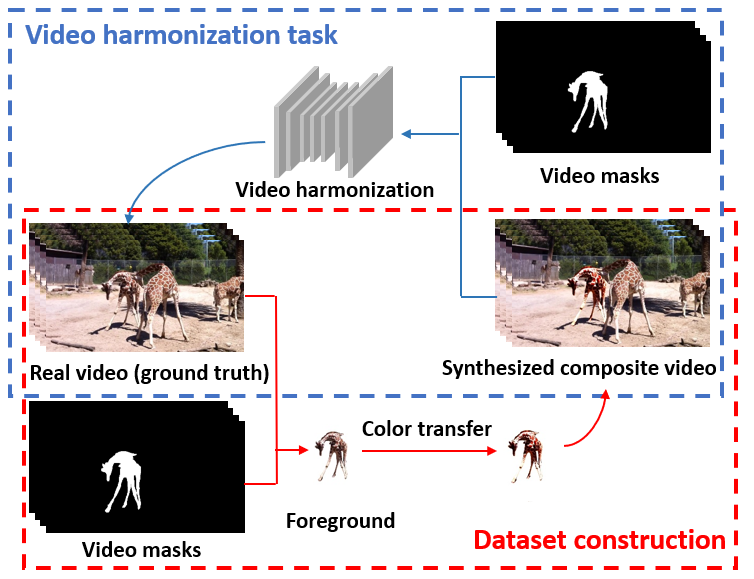}
    \caption{Illustration of video harmonization task (blue arrows) and dataset construction process (red arrows).}
    \label{fig:dataset_construction}
\end{figure}

As a closely related task, image harmonization has attracted growing research interest. 
Recently, several deep learning based image harmonization methods \cite{2020DoveNet,2020Improving,guo2021intrinsic,2021Bargainnet,2020Foreground,ling2021region} have been proposed. They changed the foreground style to be harmonious with the background using deep learning techniques. 
However, directly applying them to video harmonization by harmonizing each frame separately will cause flickering artifacts \cite{2019Temporally}. 
So \cite{2019Temporally} considered the temporal consistency between adjacent frames and proposed an end-to-end network to harmonize the composite frames.

Training deep video harmonization network requires abundant pairs of composite videos and their ground-truth harmonized videos, but manually editing composite videos to obtain their harmonized videos is tedious and expensive. 
Therefore, \cite{2019Temporally} adopted an inverse approach
by applying the traditional color transfer method \cite{946629} to the foreground of the real image to make it incompatible with the background, leading to the synthetic composite image. Then, they applied affine transformation to foreground and background to simulate the motion between adjacent frames, through which synthetic composite video (\emph{resp.}, ground-truth video) are created based on synthetic composite image (\emph{resp.}, real image). Nevertheless, there is a huge gap between the simulated movement and the complex movement in realistic videos. Moreover, their dataset is not publicly available. 
Different from \cite{2019Temporally}, we create synthetic composite videos based on real video without sacrificing realistic motion. Specifically, we apply color transfer based on lookup table (LUT)  to the foregrounds of all frames and the details will be introduced in Section \ref{dataset_construction}. We construct our video harmonization dataset named HYouTube based on YouTube-VOS 2018 \cite{xu2018YouTube}, leading to 3194 pairs of synthetic composite videos and real videos. 


To alleviate flickering artifacts, we also propose a novel video harmonization framework considering temporal coherence.  Previous video-related methods usually seek for spatial correspondence, \emph{i.e.}, the relative movement of pixels/regions between adjacent frames, using optical flow or attention mechanism.
However, it is time-consuming and challenging to establish the spatial correspondence accurately \cite{liu2020efficient}.
To escape from establishing spatial correspondence, 
we use color mapping consistency to enhance temporal coherence. 
In real videos, the color statistics between two adjacent frames should be close for both foreground and background regions. Therefore, we assume that the color transformations (mappings) to harmonize two adjacent composite frames should also be close.

Based on the assumption of color mapping consistency, we propose a framework consisting of an image harmonization network and a refinement module. 
Specifically, we first employ existing image harmonization network to harmonize composite frames. Then, for each composite frame, we can summarize the color transformation of its neighboring frames and compact it in a lookup table (LUT). Next, we apply the summarized LUT to this composite frame to obtain the LUT result. 
Finally, we feed the image harmonization result and the LUT result into a light-weighted refinement module to get our final result, during which the LUT result ensures color mapping consistency and helps improve temporal coherence. We name our framework as \textbf{CO}lor mapping \textbf{CO}nsistency \textbf{Net}work (CO$_2$Net). Our major contributions can be summarized as follows: 
\begin{itemize}
\item  We contribute the first public video harmonization dataset named HYouTube. 
\item To ensure temporal coherence, we propose a novel video harmonization framework named CO$_2$Net based on the assumption of color mapping consistency.
\end{itemize}

\section{Related Work}

\subsection{Image Harmonization}
Image harmonization aims to adjust the foreground appearance to match the background appearance. 
Recently, deep learning based image harmonization methods~\cite{2017Deep,2020DoveNet,cong2021high} mainly concentrated on learning transformation based on image-to-image translation \cite{2016Image}. \cite{2020DoveNet} proposed a domain verification discriminator to help translate the foreground to the same domain of background. \cite{2021Bargainnet} and  \cite{ling2021region} utilized the background information as guidance to translate the foreground.
The works in \cite{2020Improving,hao2020image} leveraged various attention mechanisms to improve the performance of harmonization networks. 
\cite{guo2021intrinsic} decomposed the composite image into reflectance and illumination according to Retinex theory, and harmonized the illumination map. 

\subsection{Video Harmonization}
When applying the image harmonization methods to harmonize each composite frame individually, there will be notable flickering artifacts. 
To solve this issue, video harmonization method \cite{2019Temporally} harmonized each composite frame and leveraged optical flow to make the aligned harmonized results of adjacent frames consistent. Different from this method, we leverage color mapping consistency instead of spatial correspondence to ensure the temporal coherence of a harmonized video.

\subsection{Image and Video Harmonization Datasets}
Since it is very difficult to obtain pairs of composite images and their ground-truth harmonized images, pioneering works~\cite{2017Deep,2020DoveNet} adopted an inverse approach by translating real images to synthetic composite images. The first public image harmonization dataset is iHarmony4  \cite{2020DoveNet}, which consists of four subdatasets: HCOCO, HFlickr, HAdobe5k, Hday2night. 

For video harmonization dataset, similar to~\cite{2020DoveNet}, \cite{2019Temporally} first converted real images to synthetic composite images using color transfer method. Then, affine transformation is applied to the foregrounds/backgrounds in real images and synthetic composite images synchronously to simulate the movement between adjacent frames, leading to pairs of synthetic composite videos and harmonized videos. However, the simulated movement is dramatically different from realistic motion. In contrast, we construct a video harmonization dataset by translating real videos to synthetic composite videos. 

\section{Dataset Construction}
\label{dataset_construction}

We construct our dataset HYouTube based on the large-scale video object segmentation dataset YouTube-VOS 2018 \cite{xu2018YouTube}. Given real videos with object masks, we first select the videos which meet our requirements and then adjust their foregrounds to produce synthetic composite videos.

\subsection{Real Video Selection}
\label{sec:video_select}
YouTube-VOS 2018 \cite{xu2018YouTube} contains 4453 YouTube video clips and each video clip is annotated with the object masks for one or multiple objects. Each second has 6 frames with mask annotations and we only utilize these annotated frames. 
Then, for each annotated foreground object in each video clip, if there exist more than 20 consecutive frames containing this foreground object, we save the first 20 consecutive frames with the corresponding 20 foreground masks as one video sample. After that, we remove the video samples with foreground ratio (the area of foreground over the area of the whole frame) smaller than 1\% to ensure that the foreground area is in a reasonable range. After the above filtering steps, there are 3194 video samples left.

\subsection{Composite Video Generation}
Based on real video samples, we adjust the appearance of their foregrounds to make them incompatible with backgrounds, producing synthetic composite videos. 
We have tried different color transfer methods and 3D color lookup table (LUT)  to adjust the foreground appearance. The color transfer methods \cite{2010Multi,2012RUSHMEIER,2007Using,zhu2015learning} need a reference image and adjust the source image appearance based on the reference image appearance, while LUT \cite{Mese2001Look,2011Medical} is a simple array indexing operation to realize color mapping. The details of LUT will be introduced in Section \ref{sec:Lut}. We observe that applying color transfer methods requires carefully picking reference images, otherwise the transferred foreground may have obvious artifacts or look unrealistic. Thus, we employ LUT to adjust the foreground appearance for convenience. Since one LUT corresponds to one type of color transfer, we can ensure the diversity of the composite videos by applying different LUTs to video samples. We collect more than 400 LUTs from the Internet and select 100 candidate LUTs among them with the largest mutual difference (see Supplementary).

The process of generating composite video samples is illustrated in Figure~\ref{fig:dataset_construction}. Given a video sample, we first select an LUT from 100 candidate LUTs randomly to transfer the foreground of each frame. The transferred foregrounds and the original backgrounds form the composite frames, and the composite frames form the composite video samples. Following \cite{2020DoveNet}, we set some rules to filter out unqualified composite video samples, which can be found in Supplementary.
We name our constructed video harmonization dataset as HYouTube, which includes 3194 pairs of synthetic composite video samples and real video samples. Each video sample contains 20 consecutive frames with the foreground mask for each frame.

\begin{figure*}[t]
\begin{center}
    \includegraphics[width=0.9\textwidth]{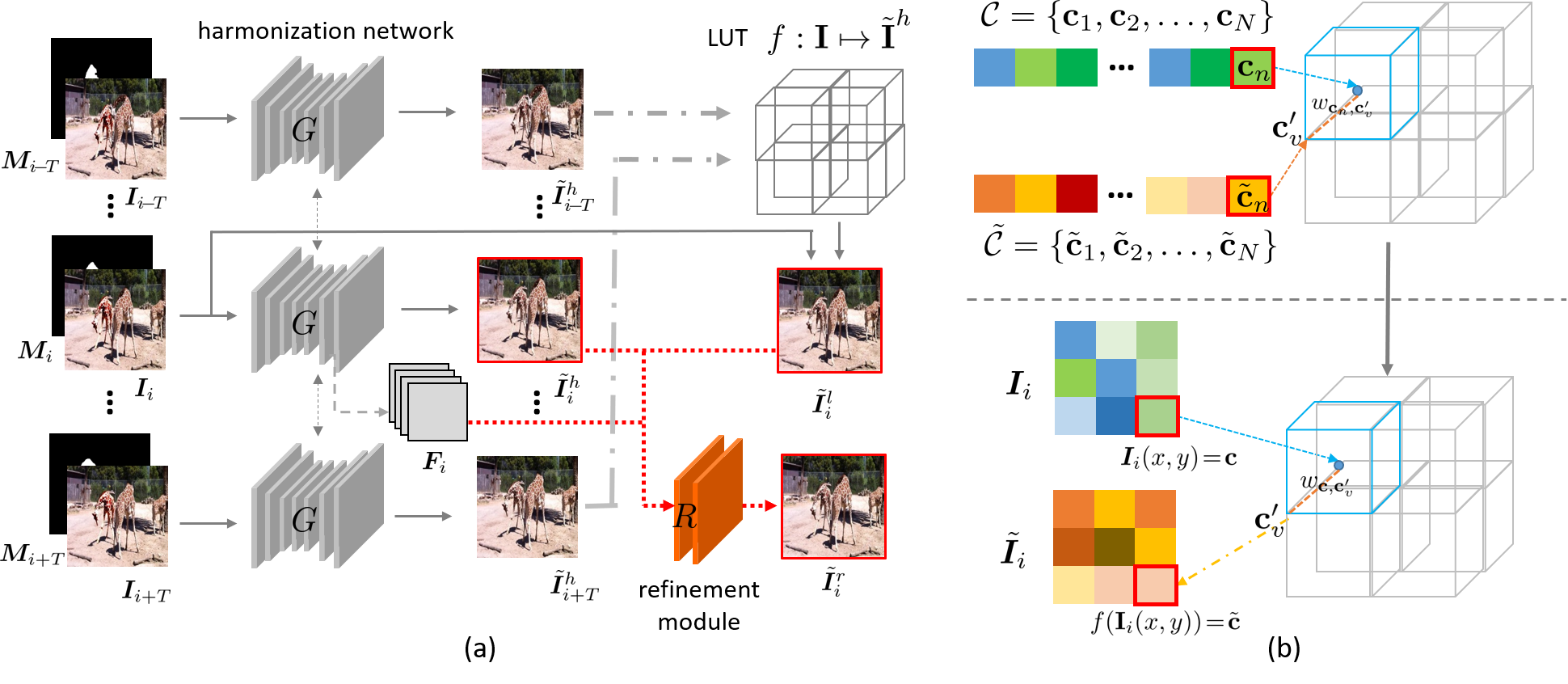}
    \caption{The subfigure (a) illustrates the whole framework. For the current frame $\mathbf{I}_i$, we can get its image harmonization result $\tilde{\mathbf{I}}_i^h$ and LUT result $\tilde{\mathbf{I}}_i^l$, which are fed into the refinement module together with decoder feature $\mathbf{F}_i$ to produce the refined result $\tilde{\mathbf{I}}_i^r$. The  subfigure (b) depicts the process of calculating LUT based on neighboring frames (top) and applying LUT to the current frame (bottom). }
    \label{fig:frame}
\end{center}
\end{figure*}

\section{Our Method}

\subsection{Framework Overview} \label{sec:frame_overview}
We propose a video harmonization framework consisting of an image harmonization network $G$ (\emph{e.g.}, \cite{ling2021region,2020Foreground}) and a refinement module $R$ as shown in Figure \ref{fig:frame}. Given an input composite video, each frame $\mathbf{I}_{i}$ concatenated with its foreground mask $\mathbf{M}_{i}$ goes through the image harmonization network to obtain the image harmonization result $\tilde{\mathbf{I}}_{i}^h$. 

Then, the refinement module takes the neighboring frames into consideration and utilizes color mapping consistency to enhance the harmonized result. Specifically, given the current frame $\mathbf{I}_{i}$, we calculate the color mapping based on its neighboring frames $\{\mathbf{I}_{i-T},\ldots, \mathbf{I}_{i-1}, \mathbf{I}_{i+1}, \ldots, \mathbf{I}_{i+T}\}$ and image harmonization results $\{\tilde{\mathbf{I}}_{i-T}^h, \ldots, \tilde{\mathbf{I}}_{i-1}^h, \tilde{\mathbf{I}}_{i+1}^h, \ldots, \tilde{\mathbf{I}}_{i+T}^h\}$, 
where $T$ is the number of neighbors on one side. We calculate the color mapping $f: \mathbf{I} \mapsto \tilde{\mathbf{I}}^h$ in the form of lookup table (LUT), which will be detailed in Section \ref{sec:Lut}.
We assume that the color mapping of the current frame should be consistent with its neighboring frames. Thus, we apply the color mapping $f$ calculated based on neighboring frames to the current frame $\mathbf{I}_{i}$ to obtain the LUT result $\tilde{\mathbf{I}}_{i}^l=f(\mathbf{I}_{i})$.
Finally, we feed LUT result $\tilde{\mathbf{I}}_{i}^l$ and image harmonization result $\tilde{\mathbf{I}}_{i}^h$ into a light-weighted refinement module to achieve the refined result $\tilde{\mathbf{I}}_i^r$.

\subsection{Color Mapping based on Lookup Table}
\label{sec:Lut}
Lookup table (LUT) records the input color and the corresponding output color, so one LUT corresponds to one color mapping function $f$.
LUT has been applied in a variety of computer vision tasks like image enhancement \cite{2011Medical,fischl1999adaptive} and image denoising \cite{2006Image,2015Randomized}.
As shown in Figure \ref{fig:frame}(b), an LUT is a 3D lattice in the RGB space and each dimension corresponds to one color channel (\emph{e.g.}, red). LUT consists of $V=(B+1)^3$ entries by uniformly discretizing the RGB color space, where $B$ is the number of bins in each dimension.
Each entry $v$ in the LUT has an indexing color $\mathbf{c}'_v=(r'_v,g'_v,b'_v)$ and its corresponding output color $\tilde{\mathbf{c}}'_v=(\tilde{r}'_v,\tilde{g}'_v,\tilde{b}'_v)$. The color transformation process based on LUT has two steps: \emph{look up} and \emph{trilinear interpolation}. Specifically, given a color value, we first look up its eight nearest entries in the LUT, and then interpolate its transformed value based on eight nearest entries via trilinear interpolation.

As introduced in Section~\ref{sec:frame_overview}, given the current frame $\mathbf{I}_{i}$, we tend to calculate the color mapping $f$ (an LUT) based on its neighboring frames. Then, we apply the color mapping $f$ to the current frame $\mathbf{I}_{i}$ to obtain the LUT result. Next, we will elaborate on these two stages one by one.



%

\subsubsection{Calculating LUT}

For ease of description, given the current frame $\mathbf{I}_i$, we collect the foreground pixels of its neighboring frames $\{\mathbf{I}_{i-T},\ldots, \mathbf{I}_{i-1}, \mathbf{I}_{i+1}, \ldots, \mathbf{I}_{i+T}\}$ as an array of pixels $\mathcal{C}=\{\mathbf{c}_1, \mathbf{c}_2,\ldots, \mathbf{c}_{N}\}$, in which $N$ is the total number of pixels. With their image harmonization results, we can obtain the harmonized pixels $\tilde{\mathcal{C}}=\{\tilde{\mathbf{c}}_1,\tilde{\mathbf{c}}_2,\ldots, \tilde{\mathbf{c}}_{N}\}$. Our goal is learning an LUT (color mapping $f$) which can translate $\mathcal{C}$ to $\tilde{\mathcal{C}}$. However, it is time-consuming and inconvenient to directly solve the optimization problem $\min_{f} \frac{1}{3N}\sum_{n=1}^N\|f(\mathbf{c}_n)-\tilde{\mathbf{c}}_n\|^2$. Hence, we design a heuristic approach which is effective in practice and much more efficient than direct optimization (see Section \ref{sec:LUT_calculation}). The intuitive idea of our heuristic approach is as follows: 1) for each $\mathbf{c}_n$, look up its nearest entries in the LUT and assign the harmonized value $\tilde{\mathbf{c}}_n$ to these entries; 2) for each entry in the LUT, aggregate the assigned values as the output color of this entry.

For each pixel $\mathbf{c}_n$, we calculate its similarity with each indexing color $\mathbf{c}'_v$ in the LUT. Given a pixel value $\mathbf{c}_n=(r_n,g_n,b_n)$ and an indexing color $\mathbf{c}'_v=(r'_v,g'_v,b'_v)$, the similarity between $\mathbf{c}_n$ and $\mathbf{c}_v'$ is calculated as
\begin{equation}
w_{\mathbf{c}_n,\mathbf{c}'_v} = \prod_{z\in \{r,g,b\}} max(0, 1 - \frac{\left| z_n - z'_v\right|}{d} ),
\label{eq:cal_w}
\end{equation}
in which $d=256/B$ is bin size. 
According to (\ref{eq:cal_w}), it can be seen that $\mathbf{c}_n$ only has non-zero similarities with its eight nearest entries and the similarities share the same form as the interpolation coefficients in trilinear interpolation. 
We assign the harmonized pixel $\tilde{\mathbf{c}}_n$ to the eight nearest entries (\emph{e.g.}, $\mathbf{c}'_v$) of $\mathbf{c}_n$ with different weights (\emph{e.g.}, $w_{\mathbf{c}_n,\mathbf{c}'_v}$). After traversing all $\mathbf{c}_n$, we can calculate the output value in each entry in the LUT as the weighted average of harmonized pixels:
\begin{equation} \label{eqn:LUT_assign}
\tilde{\mathbf{c}}'_v = \frac{\sum_{n=1}^{N} w_{\mathbf{c}_n,\mathbf{c}'_v} \tilde{\mathbf{c}}_n}{\sum_{n=1}^{N} w_{\mathbf{c}_n,\mathbf{c}'_v}}.
\end{equation}

One issue is that the pixels in $\mathcal{C}$ cannot cover all the entries in the LUT, so $\sum_{n=1}^{N} w_{\mathbf{c}_n,\mathbf{c}'_v}=0$ for some entries, which are referred to as null entries.

\subsubsection{Applying LUT}
\label{sec:apply_color_map}
The resultant LUT in (\ref{eqn:LUT_assign}) represents the color mapping function $f$ and we can apply $f$ to the current frame $\mathbf{I}_i$. In particular, given a foreground pixel $\mathbf{I}_i(x,y)=\mathbf{c}$, we can use trilinear interpolation based on the LUT to get its value $\tilde{\mathbf{c}}$ in the LUT result $\tilde{\mathbf{I}}^l_{i}$:
\begin{equation}
\tilde{\mathbf{I}}^l_{i}(x,y)= \tilde{\mathbf{c}} = f(\mathbf{c}) = \sum_{v=1}^{V} w_{\mathbf{c},\mathbf{c}'_v} \tilde{\mathbf{c}}'_v,
\end{equation}
in which the interpolation coefficient $w_{\mathbf{c},\mathbf{c}'_v}$ has the same form as (\ref{eq:cal_w}).
If the eight nearest entries of one pixel $\mathbf{I}_i(x,y)$ contain null entries, we ignore the null entries and normalize the remaining coefficients. If the eight nearest entries are all null entries, 
we refer to this pixel as an invalid pixel and replace  $\tilde{\mathbf{I}}^l_i(x,y)$ with $\tilde{\mathbf{I}}^h_i(x,y)$.

\subsection{Light-weighted Refinement Module}
Given the current frame $\mathbf{I}_i$, we can obtain its image harmonization result $\tilde{\mathbf{I}}^h_i$ and LUT result $\tilde{\mathbf{I}}^l_i$. Since $\tilde{\mathbf{I}}^h_i$ only focuses on the current frame without considering neighboring frames,  $\tilde{\mathbf{I}}^l_i$ can compensate $\tilde{\mathbf{I}}^h_i$ for the temporal consistency.

We design a light-weighted refinement module to generate the refined result $\tilde{\mathbf{I}}^r_i$ based on  $\tilde{\mathbf{I}}^h_i$ and  $\tilde{\mathbf{I}}^l_i$.
Besides, we conjecture that the last feature map $\mathbf{F}_{i} \in \mathbb{R}^{W\times H \times C}$ in the image harmonization network $G$  contains rich useful knowledge for harmonization. Thus, we also feed $\mathbf{F}_{i}$ into the refinement module. After concatenating $\tilde{\mathbf{I}}^h_i$, $\tilde{\mathbf{I}}^l_i$, and $\mathbf{F}_{i}$, we have the $H \times W \times (C+6)$ input for our refinement module. 
To balance the efficiency and performance, in the refinement module, we only employ two convolutional layers, with each one followed by batch normalization and ELU activation.


During training, we adopt a two-step training strategy. In the first step, we train the image harmonization network $G$ using its original loss functions, after which we can obtain $\tilde{\mathbf{I}}^h_i$ and $\tilde{\mathbf{I}}^l_i$ for each frame. In the second step, we train the refinement module $R$ with MSE loss $\mathcal{L} =  \|\tilde{\mathbf{I}}^r_i -\hat{\mathbf{I}}_i \|^2$, in which $\hat{\mathbf{I}}_i$ is the ground-truth harmonized result. 
We tried training the whole framework in an end-to-end manner, but it brings no further improvement while increasing the training difficulty. Therefore, we finally adopt the two-step training strategy.

\begin{table*}[t]
\centering
\begin{tabular}{|c|c|c|c|c|c|}
 \hline
Method             & fMSE$\downarrow$           & MSE $\downarrow$        & PSNR$\uparrow$         & fSSIM$\uparrow$          & TL$\downarrow$   \\ \hline
Composite              & 1029.50 & 151.20 & 30.14 & 0.7197 &                 2.5315   \\   \hline
DoveNet  \cite{2020DoveNet}         &     422.84        &      58.51        &      33.96        &      0.8238       &       13.8647             \\ \hline
IIH \cite{guo2021intrinsic}         &     368.92          &       47.30       &        34.25      &       0.8391      &      3.1187              \\ \hline
RainNet   \cite{ling2021region}          &    374.06           &     49.05         &     34.61         &     0.8338        &         4.4733             \\ \hline
iS$^2$AM    \cite{2020Foreground}         & 203.77 & 28.90 & 37.38 & 0.8817 &      6.4765               \\   \hline
Huang \emph{et al.} (RainNet)     & 373.17      & 43.94      & 34.63      & 0.8319   &      4.5044                \\ \hline
Huang \emph{et al.} (iS$^2$AM)     &  199.89      & 27.89      & 37.44      & 0.8821   &         6.4893             \\ \hline
Ours (RainNet) &    325.36           &      43.81        &     35.37         &       0.8534      & 4.0694   \\  \hline     
Ours (iS$^2$AM)   & 186.72 & 26.50 & 37.61 & 0.8827 &          5.1126           \\ \hline
\end{tabular}
\caption{Comparison between different harmonization methods on our HYouTube dataset. TL is short for temporal loss.}
\label{table:comparison_result}
\end{table*}

\begin{table}[t]
\centering
\begin{tabular}{|c|c|c|c|c|c|l|l|l|}
\hline
\multicolumn{1}{|c}{}&\multicolumn{2}{|c}{Base} & \multicolumn{3}{|c|}{Refinement}                                                                                         & \multicolumn{1}{l|}{\multirow{2}{*}{fMSE$\downarrow$}} &\multicolumn{1}{l|}{\multirow{2}{*}{Time(s)$\downarrow$}}  &  \multicolumn{1}{l|}{\multirow{2}{*}{TL$\downarrow$}} \\ \cline{1-6}

 &$\tilde{\mathbf{I}}^h$   &$\tilde{\mathbf{I}}^l$&\multicolumn{1}{|c|}{$\tilde{\mathbf{I}}^h$} & \multicolumn{1}{c|}{$\tilde{\mathbf{I}}^l$} & \multicolumn{1}{c|}{$\mathbf{F}$} & \multicolumn{1}{l|}{}    &  \multicolumn{1}{l|}{} & \multicolumn{1}{l|}{}  \\ \hline
 
1&+ & & & & &203.77 &0.0034 &6.4765\\ \hline
2&& + & & & &190.53&0.0038&6.2246\\ \hline
3&+&+&+                                  & +                               &                                                      & 189.07    &     0.0045     & 6.1983    \\ \hline
4&+&+&+   &  & +  & 202.89           &     0.0042    & 6.4917 \\ \hline
5&+&+&   & + & +  & 188.53          &     0.0046   & 6.0732 \\ \hline
6&+&+&+  & +   & +   & 186.72       &        0.0046    & 5.1126  \\ \hline

\end{tabular}
\caption{The ablation studies of our CO$_2$Net. The first two rows only use image harmonization result $\tilde{\mathbf{I}}^h$ or LUT result $\tilde{\mathbf{I}}^l$. $\tilde{\mathbf{I}}^h$, $\tilde{\mathbf{I}}^l$, and  $\mathbf{F}$ are three types of inputs for the refinement module. TL is short for temporal loss.}
\label{tab:ablation study}
\end{table}

\begin{table}[t]
\centering
\begin{tabular}{|c|c|c|c|c|c|}
\hline
Method & $B$ & $T$ & fMSE$\downarrow$ & ME$\downarrow$ & Time(s)$\downarrow$\\ \hline
Ours & 32 & 1 &199.78 &16.74 & 0.000182\\
Ours & 32 & 8 & 190.53 &25.23  &0.000185\\
Ours & 128 & 1 &205.06 &10.83 &0.000284\\
Ours & 128 & 8 &193.21 & 18.40 &0.000288\\ \hline
Op & 32 & 1 & 278.17 & 9.14 &  0.085\\
Op & 32 & 8 & 205.23 &17.85 & 0.155\\
Op & 128 & 1 & 919.07 & 1.95 &0.172\\
Op & 128 & 8 & 219.15 & 10.03 & 0.268\\ \hline
\end{tabular}
\caption{Comparison of two ways to calculate LUT. ``Ours" is our way and ``Op" is direct optimization. ME is short for mapping error.}
\label{tab:lut_calculation}
\end{table}

\section{Experiments}
\subsection{Dataset Statistics}
We conduct experiments on our constructed HYouTube dataset, which contains 3194 pairs of synthetic composite video samples and real video samples. We split our dataset into 2558 video samples in the training set and 636 video samples in the test set, in which the video samples created by adjusting different foregrounds in the same video are not allowed to appear in both training set and test set. More dataset statistics (\emph{e.g.}, the number of video samples per LUT) can be found in Supplementary.

\subsection{Implementation Details}
Our framework can accommodate an arbitrary image harmonization network $G$. We adopt two backbones iS$^2$AM  \cite{2020Foreground} and RainNet \cite{ling2021region} considering their simplicity and effectiveness. We set the number of neighboring frames $T=8$ by default. The number of bins is set as $B=32$, which is practically used in image processing. The effect of $T$ and $B$ will be discussed in Supplementary. 

We conduct all experiments using Pytorch. We train our model on a single GTX 3090 GPU for 120 epochs 
using Adam optimizer with $\beta_1$ = 0.9,  $\beta_2$ = 0.999 and $\epsilon$ = $10^{-8}$. The initial learning rate is $10^{-3}$. The batch size is set to 32 for training process.
We resize composite frames to 256$\times$256 during training and testing. The random seed is set as 5. 
When training the refinement module, for the images without adequate neighboring frames, we perform padding by replicating the first (\emph{resp.}, last) frame at the beginning (\emph{resp.}, end) of the video sample. 

We compare different methods from two perspectives: harmonization performance and temporal consistency. For harmonization performance, we adopt MSE, fMSE, PSNR following \cite{2020DoveNet} and fSSIM following \cite{guo2021intrinsic}, in which fMSE (\emph{resp.}, fSSIM) means only calculating the MSE (\emph{resp.}, SSIM) within the foreground region. For temporal consistency,  we adopt Temporal Loss (TL) following \cite{2019Temporally}. Specifically, we extract optical flows between adjacent frames and propagate the harmonized result of the previous frame to the next frame via optical flows, after which the difference between the propagated result and the original result of the next frame is calculated. More details of TL are left to Supplementary.

\begin{figure*}[h]
    \centering
    \includegraphics[width=0.99\textwidth]{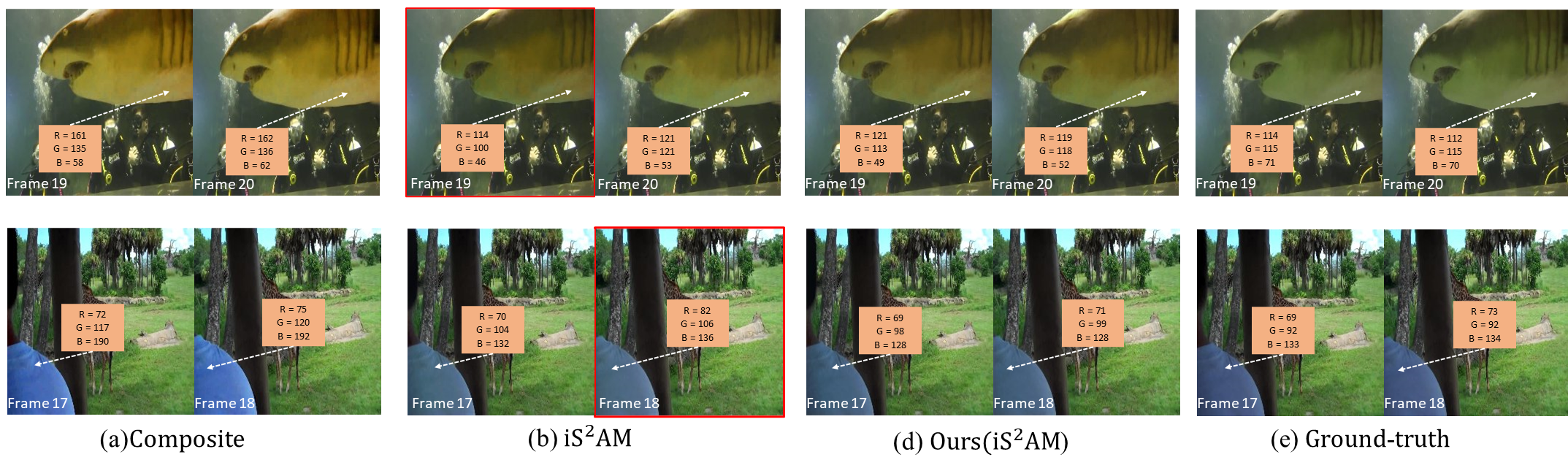}
    \caption{The qualitative comparison between iS$^2$AM and Ours (iS$^2$AM) on two adjacent frames. The frames with red borders are inconsistent with neighboring frames, which cause flickering artifacts. We also show the RGB values of temporally identical pixels in  two adjacent frames.}
    \label{fig:qualitative}
\end{figure*}

\subsection{Comparison with Existing Methods} 
We compare our CO$_2$Net with two groups of baselines: image harmonization method and video harmonization method. For the first group, we compare with existing image harmonization methods iS$^2$AM \cite{2020Foreground}, RainNet \cite{ling2021region}, DoveNet  \cite{2020DoveNet}, and 
Intrinsic Image Harmonization (IIH) \cite{guo2021intrinsic}, which harmonize each individual frame separately. For the second group, the only existing video harmonization method is   \cite{2019Temporally}. For fair comparison with our model, we also use iS$^2$AM  \cite{2020Foreground} and RainNet \cite{ling2021region} as image harmonization network for \cite{2019Temporally}. 

The experimental results are summarized in Table \ref{table:comparison_result}. Among the image harmonization methods, iS$^2$AM achieves the best results. 
When using iS$^2$AM or RainNet as the image harmonization network, \cite{2019Temporally} and our method can both improve the harmonization performance and the temporal consistency, which demonstrates the effectiveness of the temporal consistency loss in \cite{2019Temporally} and utilizing color mapping consistency in our method. 
Besides, our framework outperforms all baseline methods including \cite{2019Temporally}, which can be explained as follows.  \cite{2019Temporally} extracts optical flows to establish the spatial correspondences between adjacent frames, which is time-consuming and the established spatial correspondences are often inaccurate \cite{liu2020efficient}. In contrast, we leverage color mapping consistency to avoid establishing spatial correspondences, which is both efficient and effective. 

Following \cite{2019Temporally}, we conduct user study to evaluate the harmonized videos from two aspects: realism and temporal consistency. For each test video sample, we ask 20 users to rank the results of iS$^2$AM, Huang \emph{et al.} (iS$^2$AM), and Ours(iS$^2$AM), after which
the Plackett-Luce scores are calculated for three methods. 
The details of user study are left to Supplementary. 

\subsection{Ablation Studies}
In this section, we conduct ablation studies to analyze each component in our framework and each type of input in our refinement module in Table \ref{tab:ablation study}.
First, we report the image harmonization result in row 1 and the LUT result in row 2. Then, we add the refinement module and explore different types of inputs. 
By comparing row 3-5 with row 6, we observe that all types of inputs are essential for the refinement module. 
By comparing the time in row 1, 2, 6, it can be seen that color mapping based on LUT only costs negligible time (time in row 2 contains the time of image harmonization network) and the refinement module is also very efficient. 


\subsection{LUT Calculation}
\label{sec:LUT_calculation}
As mentioned in Section \ref{sec:Lut}, instead of directly solving the optimization problem, we design a heuristic approach to calculate the LUT, which empirically works well. We compare our approach (``Ours") with direct optimization (``Op") in different settings (\emph{e.g.}, $B$, $T$). For direct optimization, we use gradient descent to optimize the LUT by minimizing the mapping error  $\min_{f} \frac{1}{3N}\sum_{n=1}^N\|f(\mathbf{c}_n)-\tilde{\mathbf{c}}_n\|^2$. 
We report the average mapping error (ME), fMSE of LUT result, and the average time of calculating LUT by traversing all frames. The results are summarized in Table \ref{tab:lut_calculation}. Our method does not need to solve an optimization problem and works faster than ``Op".

For both ``Ours" and ``Op", the mapping error becomes smaller when $B$ increases because of the higher precision of LUT. The mapping error becomes larger when $T$ increases, because one color value may be associated with multiple harmonized color values. 

For direct optimization ``Op", we can treat $\{\mathcal{C},\tilde{\mathcal{C}}\}$ as the training set and the current frame as test set. The error on the training (\emph{resp.}, test) set is ME (\emph{resp.}, fMSE).  When $B$ increases from 32 to 128, ME decreases but fMSE increases, probably due to the overfitting issue. Similarly, larger $T$ means larger training set and leads to better generalization ability. 
``Ours" shows the same tendency. One reason is that larger $B$ and smaller $T$ will produce more invalid pixels, resulting in worse fMSE. 

\subsection{Qualitative Comparison}
Following \cite{2019Temporally}, we show two adjacent frames of the composite video, harmonized results by iS$^2$AM and Ours (iS$^2$AM), and ground-truth video in Figure \ref{fig:qualitative}. It shows that our method can produce visually appealing harmonized results which are closer to the ground-truth. We also report the RGB values of two temporally identical pixels in two frames to show the temporal consistency. We can see the two pixels in iS$^2$AM have a large color discrepancy without considering temporal consistency, while Ours (iS$^2$AM) can obtain temporally coherent results. More results are shown in Supplementary. 

\subsection{Evaluation on Real Composite Videos}
The input composite videos used in previous sections are synthetic composite videos, which may have a domain gap with real composite videos. To synthesize real composite videos, we first collect 30 video foregrounds with masks from a video matting dataset \cite{sun2021deep} as well as 30 video backgrounds from Vimeo-90k Dataset \cite{2019Video} and Internet. Then, we create composite videos via copy-and-paste and select 100 composite videos which look reasonable \emph{w.r.t.} foreground placement but inharmonious \emph{w.r.t.} color/illumination. The harmonized results and user study results are left to Supplementary. 
\subsection{Hyper-parameter Analyses and Limitations}
We investigate the impact of two hyper-parameters: the number of neighboring frames $T$ and the number of bins $B$ in the LUT.  Additionally, we discuss the limitation of our method. The above results are left to Supplementary.

\section{Conclusion}
In this paper, we have contributed a new video harmonization dataset HYouTube which consists of pairs of synthetic composite videos and ground-truth real videos. We have also designed a novel framework CO$_2$Net based on color mapping consistency. Extensive experiments on our HYouTube dataset and real composite videos have demonstrated the effectiveness of our proposed framework.

\section*{Acknowledgement}
The work is supported by the National Key R\&D Program of China (2018AAA0100704), the Shanghai Municipal Science and Technology Major Project, China (2021SHZDZX0102), National Natural Science Foundation of China (Grant No.61902247). 

\bibliographystyle{named}
\bibliography{ijcai22}

\end{document}


\maketitle

In this document, we provide additional materials to support our main submission. We describe the filtering rules of composite video sample in Section \ref{sec:filter_rules} and list some dataset statistics in Section~\ref{sec:dataset_stat}.  In Section \ref{sec:tl}, we show the detailed calculation process of temporal loss (TL). Then, we show more qualitative results and user study on HYouTube dataset (\emph{resp.}, real composite videos) in Section~\ref{sect:us} (\emph{resp.}, Section \ref{sec:Evaluation}). In Section \ref{sec:bins}, we discuss the influence of hyper-parameters of LUT. Finally, we discuss the limitation of our method in Section \ref{sec:limitation}.

\section{Filtering Rules of Composite Video Sample }\label{sec:filter_rules}
Following \cite{2020DoveNet}, we set some rules to filter out unqualified composite video samples: 1) The transferred foreground should be obviously incompatible with the background; 2) Although the transferred foreground looks incompatible with the background, the transferred foreground itself should look realistic; 3) The albedo of the foreground should remain the same after color transfer. For example, transferring a red car to a blue car is not meaningful for image harmonization~\cite{2020DoveNet}. Given a real video sample, if the obtained composite video sample after applying one LUT does not satisfy the above criteria, we will randomly choose another LUT again and repeat the transfer process until the obtained composite video sample satisfies the above criteria.

\section{Dataset Statistics}
\label{sec:dataset_stat}
We construct our HYouTube dataset based on YouTube-VOS 2018 \cite{xu2018YouTube}, so some basic statistics can be found in \cite{xu2018YouTube}. Our HYouTube dataset contains 3194 pairs of synthetic composite video samples and real video samples. We split our dataset into 2558 video samples in the training set and 636 video samples in the test set. We obtain these 3194 pairs of synthetic composite video samples by performing color transfer based on lookup table (LUT). 

Firstly, we collect more than 400 LUTs from the Internet. Secondly, we calculate their pairwise differences. Specifically, we sample 1000 real video frames. For each real frame, we apply all the collected LUTs to transfer its foreground to obtain composite frames and calculate fMSE between every two composite images as the pairwise difference between two LUTs. We average the pairwise differences over all 1000 real video frames as the final pairwise difference between two LUTs. Finally, we select 100 mutually different LUTs in an iterative approach to enlarge the diversity of synthesized composite videos.
In particular, in each iteration, we find two closest LUTs from the remaining LUTs and remove one of them. This step is repeated until there are 100 LUTs left. Thus, we get 100 candidate LUTs with the largest mutual difference, which are used to transfer the foregrounds of real video samples. 

\begin{figure}[t]
\centering    
\subfigure
{
	\begin{minipage}[t]{0.45\linewidth}
	\centering          
	\includegraphics[scale=0.26]{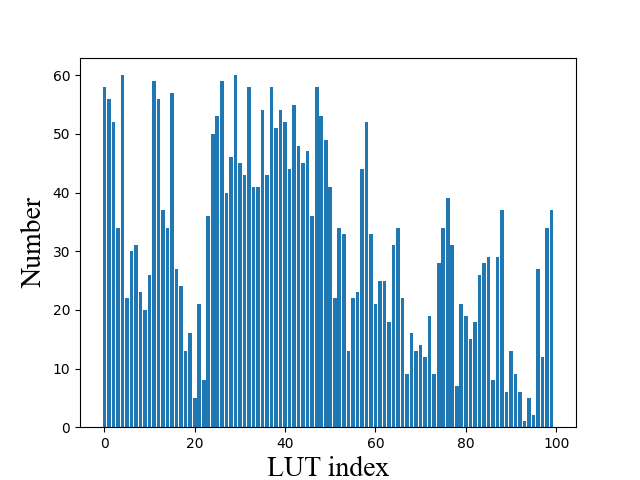}   
	\end{minipage}
}
\subfigure
{
	\begin{minipage}[t]{0.45\linewidth}
	\centering      
	\includegraphics[scale=0.26]{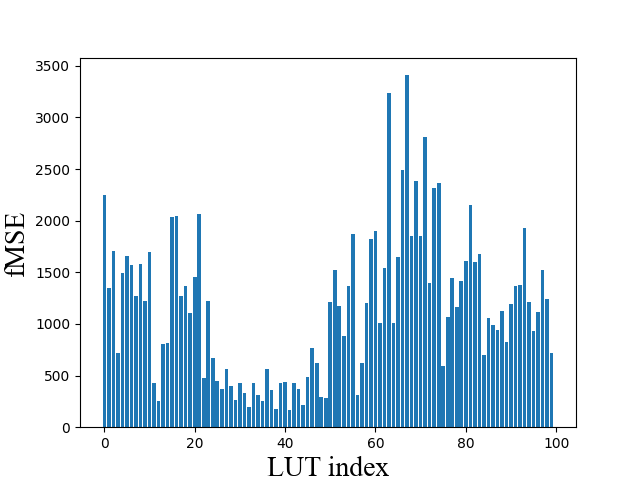}   
	\end{minipage}
}
 
\caption{The left subfigure summarizes the numbers of composite video samples created using different LUTs. The right subfigure summarizes the average fMSE between composite video samples and ground-truth video samples for different LUTs.} 
\label{fig:lut_}  
\end{figure}

\begin{figure*}[t]
    \centering
    \includegraphics[width=0.90\textwidth]{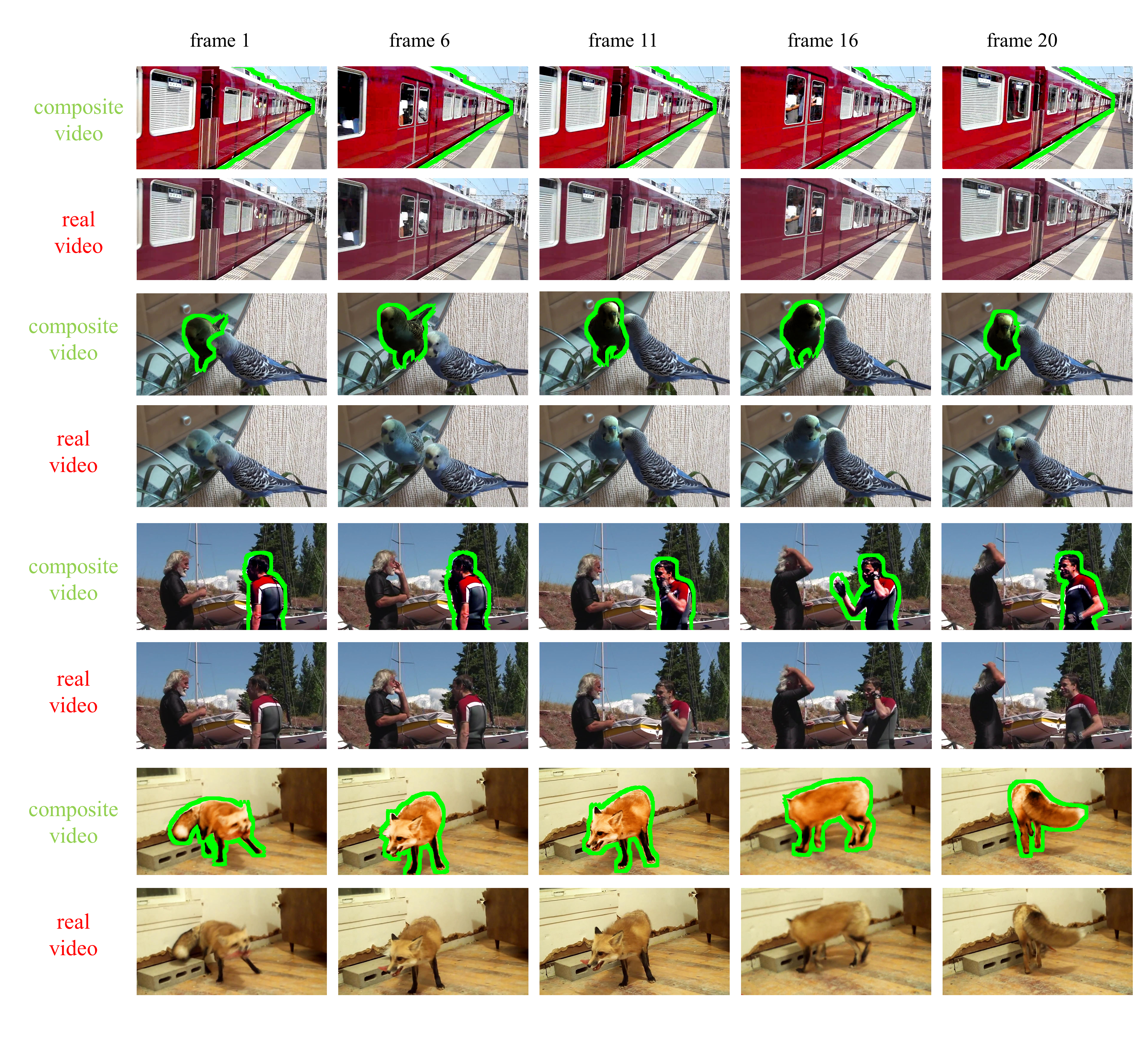}
    \caption{Some example pairs of composite video samples and real video samples. Odd rows are composite video samples and even rows are real video samples. The foregrounds are highlighted with green outlines.}
    \label{fig:dataset_samples}
\end{figure*}

The numbers of composite video samples created using different LUTs are shown in the left subfigure in Figure \ref{fig:lut_}. We can see that all 100 LUTs have been used, but some LUTs are more frequently used because they are suitable for more real video samples. The average fMSE between composite video samples and ground-truth video samples for different LUTs is shown in the right subfigure in Figure \ref{fig:lut_}. It can be seen that the average fMSE using different LUTs is quite different, which reflects the diversity of different LUTs to some extent. Finally, we show some example pairs of composite video samples and real video samples in our HYouTube dataset in Figure \ref{fig:dataset_samples}. 

\begin{figure}[t]
    \centering
    \includegraphics[width=0.5\textwidth]{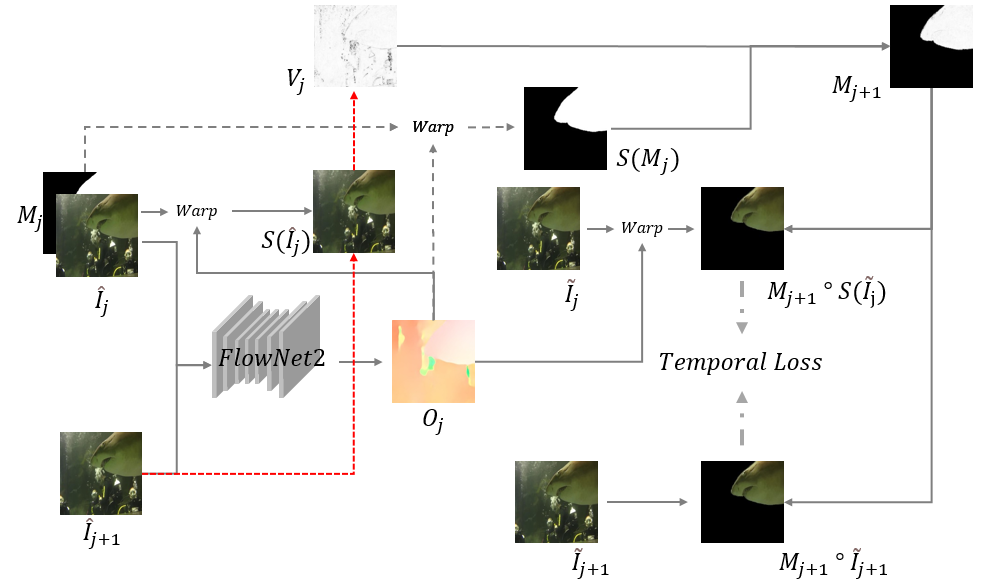}
    \caption{Illustration of obtaining the foreground mask $\mathbf{M}_{j+1}$ for the next unannotated frame and computing temporal loss (TL).}
    \label{fig:tl}
\end{figure}

\begin{figure*}[t]
    \centering
    \includegraphics[width=0.99\textwidth]{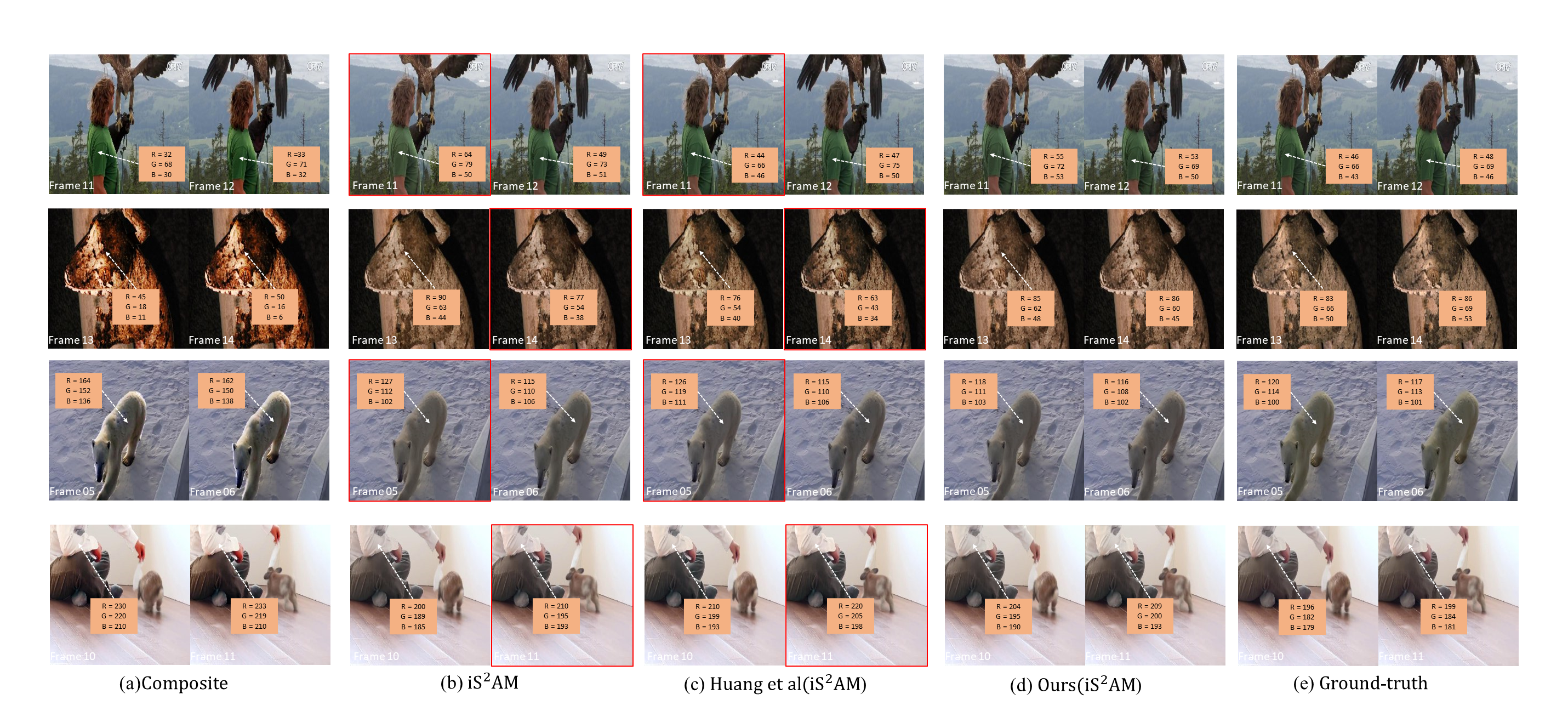}
    \caption{Qualitative result of comparison between Ours(iS$^2$AM), Huang \emph{et al.}(iS$^2$AM), and iS$^2$AM on HYouTube dataset. The frames with red borders are not consistent with their neighboring frames, which may cause flickering artifacts. We also show the RGB values of temporally identical pixels in  two adjacent frames. }
    \label{fig:qualitative-hyoutube}
\end{figure*}

\begin{figure*}[h]
    \centering
    \includegraphics[width=0.9\textwidth]{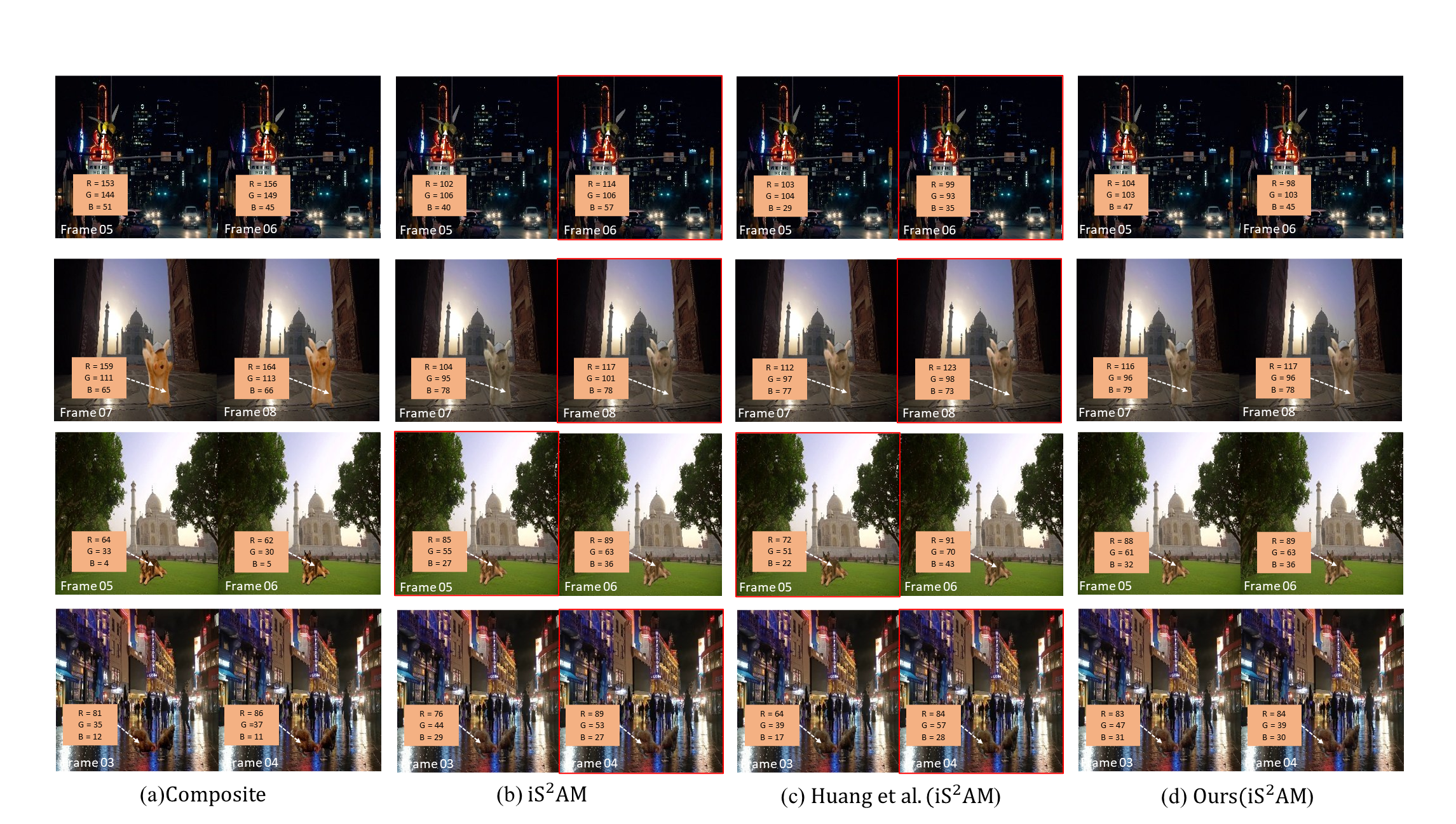}
    \caption{Qualitative result of comparison between Ours(iS$^2$AM), Huang \emph{et al.}(iS$^2$AM), and iS$^2$AM on real composite videos. The frames with red borders are not consistent with their neighboring frames, which may cause flickering artifacts. We also show the RGB values of temporally identical pixels in  two adjacent frames.}
    \label{fig:qualitative-real}
\end{figure*}

\begin{table}[t]
\centering
\begin{tabular}{|c|c|c|c|}
\hline
$T$ & fMSE$\downarrow$ & Time(s)$\downarrow$ & invalid ratio$\downarrow$  \\ \hline
0   & 203.77     & - & - \\ \hline
2            & 196.01    & 0.00025 &0.00021\\ \hline
4            & 193.10   & 0.00030 &0.00011\\ \hline
8            & 190.53   & 0.00039 &0.00006\\ \hline
12           & 189.32   & 0.00048 &0.00005\\ \hline
\end{tabular}
\caption{Performance evaluation when using different numbers of neighboring frames $T$. We report the image harmonization result for $T=0$. Invalid ratio is the average ratio of invalid pixels.}
\label{tab:parameter_T}
\end{table}

\begin{table}[t]
\centering
\begin{tabular}{|c|c|c|c|}
\hline
$B$ & fMSE$\downarrow$ & Time(s)$\downarrow$ &  invalid ratio$\downarrow$\\ \hline
16  & 195.12   & 0.00039 &0.000004\\ \hline
32  &  190.53   & 0.00039    & 0.00006\\ \hline
64  &  190.26  & 0.00040 &0.00070\\ \hline
128 & 193.21  & 0.00048 &0.0055\\ \hline
\end{tabular}
\caption{Performance evaluation when using different numbers of bins $B$ in the LUT. Invalid ratio is the average ratio of invalid pixels.}
\label{tab:parameter_B}
\end{table}


\section{Definition of Temporal Loss (TL)}
\label{sec:tl}
Following \cite{2019Temporally}, we compute temporal loss to measure the temporal consistency of harmonized videos.  We firstly compute the optical flow $\mathbf{O}_{i}$ between two adjacent ground-truth frames $\hat{\mathbf{I}}_i$ and $\hat{\mathbf{I}}_{i+1}$ by FlowNet2 \cite{2017Flownet}. After getting two harmonized frames $\tilde{\mathbf{I}}_i$ and $\tilde{\mathbf{I}}_{i+1}$, we warp the current harmonized frame $\tilde{\mathbf{I}}_i$ by optical flow $\mathbf{O}_{i}$ to obtain $S(\tilde{\mathbf{I}}_i)$. 
Then, we calculate the temporal loss (TL) within the foreground region as in \cite{2019Temporally}:
\begin{equation}
L_{T_i} = \frac{1}{3\|\mathbf{M}_{i+1}\|_1} \|\mathbf{M}_{i+1}  \circ S(\tilde{\mathbf{I}}_i) -\mathbf{M}_{i+1} \circ \tilde{\mathbf{I}}_{i+1} \|^2,
\label{eq:cal_tl}
\end{equation}
in which $\mathbf{M}_{i+1}$ is the foreground mask of the $(i+1)$-th frame, $\|\mathbf{M}_{i+1}\|_1$ is the number of foreground pixels, and $\circ$ means elementwise product. 

One practical problem is that for the video clips in YouTube-VOS 2018 dataset \cite{xu2018YouTube}, each second has 6 frames with mask annotations and 24 frames without mask annotations. We only utilize these annotated frames to build our dataset and thus the time interval between two adjacent annotated frames is 167ms. Due to the large time interval, the optical flows between two adjacent frames extracted by FlowNet2 \cite{2017Flownet} are very noisy. To shorten the time interval, we calculate temporal loss between each annotated frame and its next unannotated frame with the time interval being 33ms. 

To avoid ambiguity, we use $j$ and $j+1$ to indicate the indices of these two adjacent frames. 
The unannotated frames do not have foreground masks, so we cannot directly perform harmonization on the $(j+1)$-th composite frame. Therefore, we have to generate the foreground mask for the $(j+1)$-th composite frame by warping the current foreground mask $\bm{M}_j$ according to optical flow $\mathbf{O}_{j}$. Considering the potential occlusion which could influence the mask, we follow the same way as \cite{liu2020efficient} to generate an occlusion mask $\mathbf{V}_j$ by
\begin{equation}
\mathbf{V}_{j} =  exp(-\|\hat{\mathbf{I}}_{j+1} - \mathbf{S}( \hat{\mathbf{I}}_{j})\|^2).
\label{eq:cal_v}
\end{equation}
Then, we get the mask $\mathbf{M}_{j+1}$ by
\begin{equation}
\mathbf{M}_{j+1} =  \mathbf{V}_{j} \circ S(\mathbf{M}_{j}).
\label{eq:cal_mask}
\end{equation}
Nevertheless, we observe that the extracted optical flows are still very inaccurate. Thus, we select the adjacent frames with relatively accurate optical flows from test videos, according to the temporal loss of ground-truth frames (ground-truth TL). Specifically, we select 78 pairs of adjacent frames whose ground-truth TL is smaller than a threshold (set as 4.5 in our experiments). For evaluation, we calculate temporal loss for the harmonized results of these selected pairs. 


\begin{table}[t]
\centering
\begin{tabular}{|c|c|c|c|c|}
\hline
\multicolumn{1}{|c}{}&\multicolumn{2}{|c}{HYouTube}&\multicolumn{2}{|c|}{Real Composite}\\
\hline 
Method & R$\uparrow$ & T$\uparrow$ & R$\uparrow$ & T$\uparrow$\\ \hline
Ours(iS$^2$AM)  & 0.433  & 0.495&0.408& 0.525\\ \hline
Huang \emph{et al.}(iS$^2$AM)  & 0.341  & 0.291 &0.338&0.261\\ \hline
iS$^2$AM  & 0.226  & 0.214 &0.254&0.214\\ \hline
\end{tabular}
\caption{User study on HYouTube  and real composite videos from the perspectives of realism (R) and temporal consistency (T). We report the Plackett-Luce model scores of different methods.}
\label{tab:user_study}
\end{table}

\section{Evaluation on HYouTube Dataset}
\label{sect:us}
Following \cite{2019Temporally}, we conduct user study to evaluate the harmonized videos. We randomly sample 50 test videos and ask 20 users to rank the harmonized results of three methods: Ours (iS$^2$AM), Huang \emph{et al.} (iS$^2$AM), and iS$^2$AM from two aspects: realism and temporal consistency. For each aspect, we can get 1000 ranking lists of three methods, based on which the Plackett-Luce model scores \cite{2009Bayesian} of different methods are calculated. The Plackett-Luce model score represents the probability of ranking first among all models. 
Based on Table \ref{tab:user_study}, our method achieves the highest Plackett-Luce model scores in both aspects, which shows that our method can generate more visually appealing and temporally coherent results. 

To supplement Figure 3 in the main paper, we show more qualitative comparison in Figure \ref{fig:qualitative-hyoutube} to demonstrate the superiority of our method compared with iS$^2$AM and Huang \emph{et al.} (iS$^2$AM). 
We also observe iS$^2$AM \cite{2020Foreground} could produce harmonized results inconsistent with neighboring frames without considering temporal coherence, which may cause flickering artifacts. 
\cite{2019Temporally} employs a temporal consistency loss based on optical flow, which is often inaccurate in our dataset. Hence, the guidance role of temporal consistency loss in \cite{2019Temporally} is questionable, which makes it perform as poorly as iS$^2$AM in some cases.

Moreover, we provide several example videos in our supplementary material, in which each video contains result of Ours (iS$^2$AM), Huang \emph{et al.} (iS$^2$AM) and iS$^2$AM \cite{2020Foreground} with composite frames, mask and ground-truth frames. 
In the videos, the frames with red borders are harmonized poorly or inconsistent with their neighboring frames.

\begin{figure*}[h]
    \centering
    \includegraphics[width=0.7\textwidth]{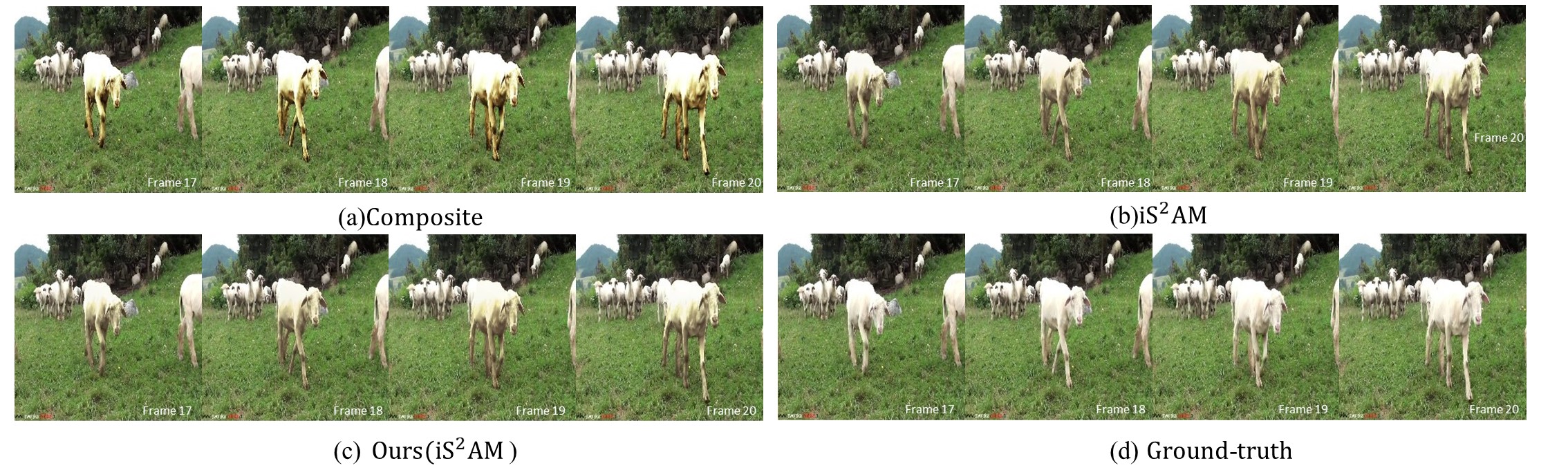}
    \caption{Illustration of failure cases of our method. When the image harmonization network (b) performs poorly on several consecutive frames, our method (c) may also have unsatisfactory performance. }
    \label{fig:wrong}
\end{figure*}

\section{Evaluation on Real Composite Videos}
\label{sec:Evaluation}
As mentioned in Section 5.7 in the main paper, we create 100 real composite videos. We apply three different methods, \emph{i.e.}, Ours (iS$^2$AM), Huang \emph{et al.} (iS$^2$AM), and iS$^2$AM to harmonize these real composite videos. Because the real composite videos do not have ground-truth, we conduct user study in the same way as in Section~\ref{sect:us} and report the Plackett-Luce model scores in Table \ref{tab:user_study}. The results demonstrate that our method can also perform better than \cite{2019Temporally} and iS$^2$AM visually in real composite videos.

We show some example composite videos and the harmonized results of three different methods in Figure \ref{fig:qualitative-real}, based on which we have similar observations as in Section~\ref{sect:us}.
Moreover, we provide several example videos including two relatively longer videos in our supplementary material, in which each video contains the result of Ours (iS$^2$AM), Huang \emph{et al.} (iS$^2$AM) and  iS$^2$AM with composite frames. 
In the videos, the frames with red borders are harmonized poorly or inconsistent with their neighboring frames.

\section{Hyper-parameter Analyses of LUT}
\label{sec:bins}



In this section, we investigate the impact of two hyper-parameters, the number of neighboring frames $T$ and the number of bins $B$ in the LUT, in Table \ref{tab:parameter_T} and Table \ref{tab:parameter_B} respectively. Because these two hyper-parameters are only related to the process of obtaining LUT result $\tilde{\mathbf{I}}^l_i$, we report harmonization performance metric fMSE for the LUT results. We also report the average time cost to obtain the LUT results (calculating LUT and applying LUT). We also report invalid ratio, which means the average ratio of invalid pixels when applying the obtained LUT to each frame (see Section 4.2 in the main paper). 


We observe that it is very efficient to obtain the LUT results. As shown in Table \ref{tab:parameter_T}, the  LUT results get better when $T$ increases, since more neighbors could offer more sufficient temporal information. 
When $T\geq 8$, the LUT results become stable, so we set $T=8$ by default. 
The invalid ratio is overall small. It decreases when $T$ increases because more color values are covered by the obtained LUT.

In the main paper, we set the number of bins in the LUT $B=32$, which is a common practice in the field of image processing. Here, we vary $B$ in $\{16, 32, 64, 128\}$ and report the results in Table~\ref{tab:parameter_B}. We report fMSE for the LUT results, as well as the average time cost to obtain the LUT results (calculating LUT and applying LUT). We also report invalid ratio, which means the average ratio of invalid pixels when applying the obtained LUT to each frame. 

We observe that fMSE first decreases and then increases when $B$ increases. The LUT with larger $B$ offers higher precision, so fMSE increases at first. However, larger $B$ also requires more color values to cover all entries, which causes the increasing invalid ratio. Since the invalid pixels are substituted with the image harmonization results and the LUT results are generally better than the harmonization results (see Table 2 in the main paper), so overlarge invalid ratio fMSE will lead to worse fMSE. 
For the time cost, $B$ determines the number of entries $V$ and mainly influences the step of Equation 2  in the main paper. This step is relatively efficient and thus the extra time cost brought by larger $B$ is negligible.

\section{Limitation}
\label{sec:limitation}
In our method, we leverage the information in neighboring frames to refine the harmonized result of current frame based on the assumption of color mapping consistency. However, as shown in Figure \ref{fig:wrong}, when the image harmonization network performs poorly on most frames, the information in neighboring frames may not help improve the harmonized result of current frame.




\bibliographystyle{named}
\bibliography{ijcai22}